\documentclass[twoside,11pt]{article}

\usepackage{blindtext}
\usepackage{hyperref}
\usepackage{dmlr2e}
\usepackage{longtable}
\usepackage{multirow}
\usepackage{booktabs}
\usepackage{csquotes}
\usepackage{xcolor}
\usepackage{enumitem}
\setlist{nosep}

\usepackage{lastpage}
\def\openreview{\url{https://openreview.net/forum?id=wGvflAGXSb}} % Insert correct link to OpenReview for camera-ready version
\begin{document}

\title{Birbal: An efficient 7B instruct-model fine-tuned with curated datasets}
\author{\name Ashvini Kumar Jindal \email ajindal@linkedin.com \\
       \addr LinkedIn AI, USA
       \AND
       \name Pawan Kumar Rajpoot \email pawan.rajpoot2411@gmail.com \\
       \addr SCB DataX, Thailand
       \AND
       \name Ankur Parikh \email ankur.parikh85@gmail.com \\
       \addr UtilizeAI Research, India
       }

% \editor{My editor}
\maketitle

\begin{abstract}%   <- trailing '%' for backward compatibility of .sty file
LLMOps incur significant costs due to hardware requirements, hindering their widespread accessibility. Additionally, a lack of transparency in model training methods and data contributes to the majority of models being non-reproducible. To tackle these challenges, the LLM Efficiency Challenge was introduced at NeurIPS Workshop\footnote{\url{https://llm-efficiency-challenge.github.io}}, aiming to adapt foundation models on a diverse set of tasks via fine-tuning on a single GPU (RTX 4090 or A100 with 40GB) within a 24-hour timeframe. In this system description paper, we introduce \textbf{Birbal}, our Mistral-7B based winning model, fine-tuned on a single RTX 4090 for 16 hours. Birbal's success lies in curating high-quality instructions covering diverse tasks, resulting in a 35\% performance improvement over second-best Qwen-14B based submission.
\end{abstract}


\section{Introduction}

Few-shot Large Language Models (LLMs) have excelled in various NLP tasks, from standardized exams \cite{openai2023gpt4, ahmed2023chatgpt, singhal2022large} to coding challenges and chatbots \cite{singhal2022large, hanschatgpt}. Typically, this involves fine-tuning an LLM with associated task(s) examples. However, the costs of fine-tuning and querying LLMs to perform new tasks are large due to the expensive and often proprietary hardware used to train and serve these models. Given these costs, access to performant LLMs has been gated, making them inaccessible to those without substantial resources.

Despite the rise of open-source LLMs like Llama-2 \cite{touvron2023llama}, Falcon \cite{almazrouei2023falcon}, Qwen \cite{qwen}, and Mistral \cite{jiang2023mistral}, the field encounters challenges in reproducibility and transparency. Many LLMs release partial artifacts, offering only final model weights or inference code, hindering comprehensive disclosure of training methodologies and impeding researchers' ability to replicate reported results. For instance, Llama \cite{touvron2023llama} discloses training data mixtures, but the absence of data processing and training code impedes full reproducibility, as observed in the case of RedPajama \cite{together2023redpajama}, an open reproduction of Llama's data.

To address the lack of transparency in model training and democratize access to cutting-edge LLMs, a LLM efficiency challenge was introduced at the NeurIPS Workshop. This challenge required participants to fine-tune an open-source foundation model on a single GPU (RTX 4090 or A100 with 40GB) within a 24-hour timeframe. In this paper, we introduce \textbf{Birbal}, our Mistral-7B based winning model, fine-tuned with high-quality instructions covering diverse tasks on a single RTX 4090 for 16 hours.

\section{LLM Efficiency Challenge}

The LLM Efficiency Challenge tasks participants with fine-tuning an open-source \enquote{base} language model on a single GPU (RTX4090 or A100 40GB) for 24 hours, exclusively using open-source data. The competition has two hardware tracks: the NVIDIA 4090 track and the NVIDIA A100 track. Accepted base models must be open and without instruction-tuning, adhering to licenses like MIT, Apache 2, BigScience RAIL, and Llama-2 Community License Agreement. Participants can use various standard autoregressive and autoencoder base models and all open-source datasets, including Databricks-Dolly-15 \cite{DatabricksBlog2023DollyV2}, OpenAssistant Conversations Dataset (oasst1) \cite{kopf2023openassistant}, The Flan Collection \cite{longpre2023flan}, AllenAI Dolma \cite{dolma}, RedPajama-Data-1T \cite{together2023redpajama}, and LIMA \cite{zhou2023lima}, are allowed, emphasizing avoidance of datasets with generated content unless explicitly permitted by the source model's license. Each team can submit three entries per track, and post-competition, winning models, code, and data must be open-sourced. The evaluation consists of four stages, where submissions are assessed on a set of tasks, and rankings are determined by the geometric mean across all evaluation tasks. Submissions below a score threshold are eliminated in stages 1 and 2. Stage-1 evaluates submissions on a subset of HELM \cite{lee2023holistic} tasks (open eval), and stage-2 assesses them on a hidden evaluation set (closed eval). In the third stage, organizers reproduce training artifacts for consistency, and in the final stage, submissions are evaluated and ranked on a subset of open and closed tasks' performance.

\section{Our Approach}
\subsection{Design Choices}
We participated in the RTX 4090 track of the competition. In this section, we outline our design choices based on the aforementioned constraints:
\begin{itemize}
    
\item \textbf{Data Sources} - Evaluation included both HELM and hidden tasks; to excel in the latter, minimize reliance on HELM-sourced data.

\item \textbf{Mistral-7B vs Qwen-14B}: Within 24GB memory budget, Mistral-7B and Qwen-14B were best performing models outperforming Llama-2 13B on several benchmarks\footnote{\url{https://huggingface.co/spaces/HuggingFaceH4/open\_llm\_leaderboard}}.

\item \textbf{High-Quality Data vs. Hardware Optimization}: We can optimize performance through kernel optimizations or prepare a high-quality dataset for fine-tuning.

\item \textbf{Dataset Curation vs Generation:} The success of recent LLM-generated datasets like Stanford Alpaca \cite{alpaca} is promising. However, only open-source base models can be used for dataset generation. We can curate a high-quality dataset from existing sources or generate from base LLMs.

% With the recent success of LLM-generated datasets like Stanford Alpaca \cite{alpaca}, it is a very promising direction. However, only open-source base models can be used to generate datasets. We can curate a high-quality dataset from existing sources or generate base LLMs.
\end{itemize}

\subsection{Strategy}
The spirit of the competition was to create a model that works well on diverse tasks. Based on this, we chose Mistral-7B base model \cite{jiang2023mistral} to fit more high-quality instructions covering multiple tasks. Due to our limited practical exposure in hardware optimization, we focused on high-quality dataset construction. Moreoever, we chose to curate existing datasets as generating datasets with a relatively small model (7B) can be tricky.
\subsection{Data Curation}
Our dataset curation methodology was geared toward obtaining various datasets spanning a broad spectrum of tasks. Given the constraints of 24GB memory and 24-hour fine-tuning limit, we determined that 200K, 400K, and 700K size datasets can be fine-tuned for three epochs, two epochs, and one epoch, respectively. Our data curation method is explained below.  Table \hyperlink{table1}1  shows a summary of the final datasets.
\begin{itemize}
\item \textbf{LIMA} \cite{zhou2023lima} – This is a set of 1,000 well-crafted prompts and responses utilized by the LIMA model. We added all these data points in our final datasets.  

\item \textbf{Open-Platypus} \cite{platypus2023} – A subset from various open datasets employed in Platypus models. We excluded 10\% of the dataset containing GPT-generated instructions to satisfy the challenge constraint. We added the remaining data points in our final datasets.

\item \textbf{Natural Instructions} \cite{naturalinstructions,supernaturalinstructions} – The Natural Instructions (NI) dataset is an extensive assemblage of over 1,600 tasks, each defined through natural language instructions, and includes crucial metadata such as task outline, domain, category, and input/output languages. We sample examples from NI dataset based on strategy described later.

\item \textbf{Other datasets} – In addition to above datasets, we randomly sampled examples from HELM training datasets: \textbf{OpenbookQA} \cite{mihaylov-etal-2018-suit}, \textbf{QUAC} \cite{choi-etal-2018-quac}, and \textbf{CNN/DailyMail} \cite{see-etal-2017-get, hermann2015teaching}. To bolster the model's mathematical reasoning capabilities, we randomly sampled examples from the \textbf{MathInstruct} \cite{yue2023mammoth} dataset, post exclusion of LLM-generated examples. \\
\end{itemize}

\noindent Our NI dataset curation process, focused on winning 200k dataset, consists of four stages:
\begin{enumerate}
\item \textbf{Tasks subset selection} - We selected a subset of 463 tasks\footnote{\url{https://github.com/Upaya07/NeurIPS-llm-efficiency-challenge/blob/main/selected_NI_tasks}} from the total pool of 1600+ tasks. Tasks featuring non-English inputs/outputs were eliminated, resulting in the exclusion of 576 tasks. We also disregarded tasks from the MMLU benchmark in the Question Answering category, as usage of the MMLU dataset was not allowed in the competition. Tasks falling under Question Generation and Question Understanding categories were excluded in favor of focusing on answer generation tasks. Tasks in the Wrong Candidate Generation and math categories were also removed. Additionally, tasks related to linguistic aspects such as PoS tagging, Keyword Tagging, Named Entity Recognition, Coreference Resolution, Word Semantics, Linguistic Probing, and Paraphrasing were filtered out, considering the inherent strength of most LLMs in linguistics and general text understanding tasks. Consequently, the chosen tasks spanned 33 categories within the NI dataset, heavily weighted toward the more prevalent categories, encompassing tasks related to Question Answering, Sentiment Analysis, Program Execution, Toxic Language Detection, and others.
\item \textbf{Task Categorization} - Each of the 463 selected tasks is manually categorized as \enquote{Exact Match} or \enquote{Generation}, based upon the output characteristics of the task.
\item \textbf{Few-Shot Inference} - For quantitative assessment of tasks performance, we performed few-shot inference on Mistral-7B base model to direct controlled generation. We used Accuracy for \enquote{Exact Match} tasks and ROUGE score for \enquote{Generation} tasks.
\item \textbf{Sampling} - For 200k dataset, We sampled 50K examples from both \enquote{Exact Match} and \enquote{Generation} tasks. For \enquote{Exact Match} tasks, first, we removed low-accuracy tasks as those tasks might be too difficult for the model to learn within our constraints. Then, we bucketed tasks based on their accuracy. Next, we sample examples from each task in a bucket. Specifically, we sample more examples from lower accuracy tasks and vice-versa. Finally, we randomly selected 50K examples from the aggregated pool. For \enquote{Generation} tasks, we bucketed examples from each task based on the ROUGE score. The buckets were [0,0.2), [0.2,0.3), [0.3,0.4), [0.4,0.5), [0.5,0.6), [0.6,0.7), and [0.7,0.8). For each task, we randomly sampled 40\% examples from the bucket [0,0.2] and 10\% examples from the remaining buckets. Finally, we randomly selected 50K examples from the aggregated pool. 
\end{enumerate}

\begin{center}
\begin{tabular}{cccc}
\textbf{Source Dataset}\hypertarget{table1} & \textbf{200K} & \textbf{400K} & \textbf{700K} \\ \hline
LIMA & 1K & 1K & 1K \\
Open-Platypus & 25K & 25K & 25K \\
NI (Exact Match) & 50K & 110K & 220K \\
NI (Generation) & 50K & 110K & 220K \\
OpenQA & 5K & 5K & 5K \\
QUAC & 10K & 10K & 10K \\
CNN/DailyMail & 15K & {28k} & {28k} \\
MathInstruct & 50K & 100K & 200K \\

\end{tabular}
\end{center}
\hspace{3ex}\textbf{Table 1: Our curated datasets\footnote{\url{https://github.com/Upaya07/NeurIPS-llm-efficiency-challenge\#birbal-models-and-datasets}}}. NI refers to Natural-Instructions dataset.

\subsection{Fine-Tuning}
Due to memory and fine-tuning time constraints, we applied 4-bit QLoRA \cite{dettmers2023qlora} to fine-tune the Mistral-7B base model. To meet the time limit, we conducted fine-tuning for $\sim$3 epochs on the 200K dataset, $\sim$2 epochs on the 400K dataset, and $\sim$1 epoch on the 700K dataset. We randomly sampled 2000 examples from fine-tuning dataset as validation set. For LoRA, we set the rank to 128 and alpha to 256. We apply LoRA to all Query, Key, and Value metrics in multi-head self-attention blocks alongside Linear layers. Following NEFTune \cite{jain2023neftune}, random noise was introduced into embeddings. We set gradient accumulation steps to 3 with the micro-batch size of 2 to simulate a larger batch size. We used paged\_adamw\_32bit optimizer with a cosine schedule with a learning rate of 2e-5. We set the decay rate to 0.01 and warmup steps to 100. Additionally, we enabled sample packing to enhance fine-tuning efficiency. All fine-tuning experiments were conducted using axolotl\footnote{\url{https://github.com/OpenAccess-AI-Collective/axolotl}} library. After fine-tuning for 24 hours, we selected a checkpoint based on minimum validation loss and used it for final submissions. We submitted three fine-tuning models on 200K, 400K, and 700K size datasets, respectively.
% \begin{figure}
% \includegraphics[width=5.2cm,scale=5,]{loss.png}
% \caption{Loss}
% \end{figure}

\section{Evaluation}
In the initial evaluation stage (Open Eval), all submissions underwent assessment on a subset of HELM tasks, featuring test examples sourced from datasets like MMLU \cite{hendryckstest2021, hendrycks2021ethics}, TruthfulQA \cite{lin-etal-2022-truthfulqa}, BBQ \cite{parrish-etal-2022-bbq}, GSM8K \cite{cobbe2021gsm8k}, and Big-bench \cite{srivastava2023beyond}. Among the 57 submissions in the 4090 track, 30 qualified for the subsequent stage based on a predefined threshold. From our three submissions (Birbal-200k, Birbal-400k, Birbal-700k), Birbal-200K secured the 20th rank with a score of 0.64, while the other two did not progress to the second evaluation stage. In the second stage, the 30 selected submissions were evaluated on hidden tasks, leading to the selection of 10 teams for the model reproducibility stage. Test examples were drawn from datasets like SAMSum \cite{gliwa-etal-2019-samsum}, Corr2cause \cite{jin2023large}, MATH \cite{hendrycksmath2021}, and ETHICS \cite{hendrycks2021ethics}. Our team achieved the 1st rank with a score of 0.660. The model was successfully reproduced in the third stage, and in the final evaluation stage, a new subset was formed from datasets in the first and second stages. The final score was computed as a weighted sum of open and closed eval scores, with 1/3 and 2/3 weights, respectively. The scores of the top 3 teams are detailed in Table \hyperlink{table2}2.

\begin{longtable}{cccccc}
% \hypertarget{table2}
\toprule
      &       &    & \multicolumn{3}{c}{Top-3 Teams\footnote{\url{https://llm-efficiency-challenge.github.io/leaderboard}} Scores} \\
\cmidrule(lr){4-6}
Dataset\hypertarget{table2}&Metric&Stage&Birbal{*}&Rank-2{†}&Rank-3{*} \\
&&&\textbf{(Ours)}&& \\ \midrule
\multirow{4}{*}{MMLU}
& {EM(Accuracy)}&Open&0.63&0.69&0.64\\
& {EM(Robustness)}&Open&0.59&0.64&0.60\\
& {EM(Fairness)}&Open&0.60&0.65&0.60\\
& {\textbf{MWR}}&Open&0.42&\textbf{0.87}&0.46\\ \midrule
\multirow{4}{*}{TruthfulQA}
&{EM(Accuracy)}&Open&0.59&0.52&0.57\\
&{EM(Robustness)}&Open&0.54&0.52&0.52\\
&{EM(Fairness)}&Open&0.49&0.44&0.46\\
&{\textbf{MWR}}&Open&\textbf{0.75}&0.28&0.56\\ \midrule
\multirow{2}{*}{BIG-bench}
&{EM(Accuracy)}&Open&0.33&0.38&0.0\\
&{\textbf{MWR}}&Open&0.75&\textbf{0.87}&0.06\\ \midrule
\multirow{2}{*}{GSM8K}
&{EM(Accuracy)}&Open&0.44&0.57&0.0\\
&{\textbf{MWR}}&Open&0.62&\textbf{0.81}&0.03\\ \midrule
\multirow{2}{*}{BBQ}
&{EM(Accuracy)}&Open&0.74&0.85&0.93\\
&{\textbf{MWR}}&Open&0.25&0.56&\textbf{0.75}\\ \midrule
\multirow{6}{*}{sam\_sum}
&{ROUGE-2}&Closed&0.13&0.03&0.10\\
&{Stereotypes(race)§}&Closed&0.67&-&0.67\\
&{Stereotypes(gender)§}&Closed&0.45&0.42&0.34\\
&{Representation(race)§}&Closed&0.46&0.62&0.38\\
&{Representation(gender)§}&Closed&0.01&0.0&0.01\\
&{\textbf{MWR}}&Closed&\textbf{0.38}&0.21&0.65\\ \midrule
\multirow{2}{*}{corr2cause}
&{EM(Accuracy)}&Closed&0.61&0.47&0.50\\
&{\textbf{MWR}}&Closed&\textbf{0.87}&0.25&0.62\\ \midrule
\multirow{2}{*}{MATH}
&{chain-of-thoughts}&Closed&0.12&0.07&0.05\\
&{\textbf{MWR}}&Closed&\textbf{0.75}&0.5&0.25\\ \midrule
\multirow{3}{*}{ethics\_j}
&{EM(Accuracy)}&Closed&0.68&0.68&0.70\\
&{EM(Robustness)}&Closed&0.64&0.66&0.65\\
&{EM(Fairness)}&Closed&0.62&0.58&0.64\\ \midrule
\multirow{3}{*}{ethics\_c}
&{EM(Accuracy)}&Closed&0.41&0.52&0.49\\
&{EM(Robustness)}&Closed&0.33&0.45&0.42\\
&{EM(Fairness)}&Closed&0.34&0.5&0.45\\ \midrule
\multirow{3}{*}{ethics\_v}
&{EM(Accuracy)}&Closed&0.89&0.77&0.74\\
&{EM(Robustness)}&Closed&0.86&0.70&0.67\\
&{EM(Fairness)}&Closed&0.86&0.69&0.69\\ \midrule
\multirow{3}{*}{ethics\_d}
&{EM(Accuracy)}&Closed&0.63&0.58&0.60\\
&{EM(Robustness)}&Closed&0.58&0.49&0.52\\
&{(Fairness)}&Closed&0.59&0.53&0.49\\ \midrule
\multirow{3}{*}{ethics\_u}
&{EM(Accuracy)}&Closed&0.72&0.55&0.56\\
&{EM(Robustness)}&Closed&0.60&0.34&0.45\\
&{EM(Fairness)}&Closed&0.64&0.40&0.52\\ \midrule
\multirow{1}{*}{ethics}
&{\textbf{MWR}}&Closed&\textbf{0.55}&0.41&0.47\\ \midrule
 \multirow{1}{*}{Open Eval Score}
&{}&&0.52&\textbf{0.63}&0.21\\
\multirow{1}{*}{Closed Eval Score}
&{}&&\textbf{0.61}&0.32&0.47\\ \bottomrule
\multirow{1}{*}{\textbf{Final Score}}
&{}&&\textbf{0.58}&0.42&0.38\\
\bottomrule
\end{longtable}%
\textbf{Table-2: Comparative Analysis of Top-3 Models' Overall Performance on Open and Closed Tasks.} \textit{Open Eval Score} and \textit{Closed Eval Score} for each submission are derived as the geometric mean of mean win rates across tasks in the \textit{Open} and \textit{Closed} evaluation stages, respectively. The \textit{Final Score} is computed as a weighted sum of the \textit{Open Eval Score} (weighted at 1/3) and \textit{Closed Eval Score} (weighted at 2/3). \textit{ethics\_justice}, \textit{ethics\_commonsense}, \textit{ethics\_virtue}, \textit{ethics\_deontology}, and \textit{ethics\_utilitarianism} are denoted as \textit{ethics\_j}, \textit{ethics\_c}, \textit{ethics\_v}, \textit{ethics\_d}, and \textit{ethics\_u}, respectively. \textit{Birbal} is fine-tunined model on 200k dataset. \textit{MWR} refers to Mean Win Rate. \textit{EM} refers to Exact Match. § refers to lower is better. (* = Mistral-7B as base model, † = Qwen-14B as base model)

\begin{longtable}{ccccccc}
% \hypertarget{table3}
\toprule
&&&& \multicolumn{3}{c}{Model Variants} \\
\cmidrule(lr){5-7}
Dataset\hypertarget{table3}&Metric&Stage&Mistral-7B&Birbal&Birbal&Birbal \\
&&&&(200k)&(400k)&(700k) \\ \midrule
\multirow{3}{*}{MMLU}
& {EM(Accuracy)}&Open&\textbf{0.64}&0.63&0.62&0.62 \\
& {EM(Robustness)}&Open&\textbf{0.60}&0.59&0.57&0.58 \\
& {EM(Fairness)}&Open&0.59&\textbf{0.60}&0.58&0.59\\ \midrule
\multirow{2}{*}{TruthfulQA}
&{EM(Accuracy)}&Open&0.56&\textbf{0.59}&0.41&0.46\\
&{EM(Robustness)}&Open&0.44&\textbf{0.54}&0.38&0.43\\
&{EM(Fairness)}&Open&0.44&\textbf{0.49}&0.34&0.39\\ \midrule
\multirow{1}{*}{BIG-bench}
&{EM(Accuracy)}&Open&0.29&0.33&\textbf{0.38}&0.37\\ \midrule
\multirow{1}{*}{GSM8K}
&{EM(Accuracy)}&Open&0.33&0.44&\textbf{0.61}&0.56\\ \midrule
\multirow{1}{*}{BBQ}
&{EM(Accuracy)}&Open&\textbf{0.80}&0.74&0.67&0.64\\ \midrule
\multirow{5}{*}{sam\_sum}
&{ROUGE-2}&Closed&0.14&0.13&\textbf{0.16}&\textbf{0.16}\\
&{Stereotypes(race)§}&Closed&-&0.67&-&-\\
&{Stereotypes(gender)§}&Closed&0.32&0.45&\textbf{0.30}&0.35\\
&{Representation(race)§}&Closed&\textbf{0.33}&0.46&\textbf{0.33}&\textbf{0.33}\\
&{Representation(gender)§}&Closed&0.05&0.01&0.01&\textbf{0.0}\\ \midrule
\multirow{1}{*}{corr2cause}
&{EM(Accuracy)}&Closed&0.46&\textbf{0.61}&0.55&0.56\\ \midrule
\multirow{1}{*}{MATH}
&{chain-of-thoughts}&Closed&0.02&\textbf{0.12}&\textbf{0.12}&\textbf{0.12}\\ \midrule
\multirow{3}{*}{ethics\_j}
&{EM(Accuracy)}&Closed&0.69&0.68&\textbf{0.71}&\textbf{0.71}\\
&{EM(Robustness)}&Closed&0.64&0.64&0.68&\textbf{0.69}\\
&{EM(Fairness)}&Closed&0.61&0.62&0.61&\textbf{0.65}\\ \midrule
\multirow{3}{*}{ethics\_c}
&{EM(Accuracy)}&Closed&\textbf{0.51}&0.41&0.40&0.43\\
&{EM(Robustness)}&Closed&0.36&0.33&0.36&\textbf{0.39}\\
&{EM(Fairness)}&Closed&\textbf{0.42}&0.34&0.36&0.39\\ \midrule
\multirow{3}{*}{ethics\_v}
&{EM(Accuracy)}&Closed&0.68&\textbf{0.89}&0.79&0.81\\
&{EM(Robustness)}&Closed&0.58&\textbf{0.86}&0.77&0.77\\
&{EM(Fairness)}&Closed&0.56&\textbf{0.86}&0.76&0.77\\ \midrule
\multirow{3}{*}{ethics\_d}
&{EM(Accuracy)}&Closed&0.64&0.63&\textbf{0.67}&0.65\\
&{EM(Robustness)}&Closed&0.51&0.58&\textbf{0.59}&\textbf{0.59}\\
&{(Fairness)}&Closed&0.54&0.59&\textbf{0.60}&\textbf{0.60}\\ \midrule
\multirow{3}{*}{ethics\_u}
&{EM(Accuracy)}&Closed&0.61&\textbf{0.72}&0.66&0.64\\
&{EM(Robustness)}&Closed&0.42&\textbf{0.60}&0.58&0.59\\
&{EM(Fairness)}&Closed&0.49&\textbf{0.64}&0.58&0.57\\
\bottomrule
\end{longtable}%

\textbf{Table-3: Overall performance of Birbal models fine-tuned on different dataset sizes vs Mistral-Base-7B on open and closed eval.} Best scores are marked in \textbf{bold}. § refers to lower is better.   \\

Though our 400k and 700k submissions could not make it to the second stage in the competition, we have benchmarked all submissions on all evaluation scenarios in Table-\hyperlink{table3}3 for detailed analysis. Birbal-200k, Birbal-400k, and Birbal-700k models were fine-tuned for 3, 2, and 1 epoch(s), respectively. There are a total of 31 evaluations: 9 Open and 22 Closed evaluations. Mistral-7B base model scored best in 3 open and 3 closed evaluations. Birbal-200k scored best in 4 open and 8 closed evaluations. Birbal-400k scored best in the 2 open and 8 closed evaluations. Bibral-700k scored best in 10 closed evaluations. During fine-tuning, adding more data points led to a drop in performance in a number of open tasks. However, performance on closed tasks improves as we scale a number of data points.

\section{Conclusion}
This paper describes fine-tuning the base Mistral-7B model on our curated subset of existing datasets on RTX 4090 (24 GB) GPU for one day. The fine-tuned model was evaluated on various tasks and outperformed other submissions by more than 35\%. % in the final evaluation score. This shows that fine-tuning a model with a high-quality dataset that covers large number of tasks is the key for generalization.
     
\impact{This work addresses the challenge of adapting an LLM with only 1 GPU (24GB or 40GB memory) within a day. So, this type of approach has the potential to make an efficient LLM fine-tuning accessible to those without substantial resources. Our open-source model was developed by fine-tuning the Mistral-7B model on a subset of datasets without alignment. So, it contains certain forms of bias (e.g., the risk of social stereotypes, discrimination and exclusion, and the risk of under-representing certain languages or domains) that are already present in the base model and source datasets.} 

\acks{We would like to thank Lambda Labs\footnote{\url{https://lambdalabs.com/}} for providing the compute resources required for post-competition analysis.}

\section{Reproducibility}
Failure to reproducibility was one of the criteria to eliminate the submissions from the competition. Our winning model was reproduced successfully. Our dataset curation mechanism, fine-tuning scripts, and models are publicly available\footnote{\url{https://github.com/Upaya07/NeurIPS-llm-efficiency-challenge}}.

\vskip 0.2in
\bibliography{sample}
% \appendix
% \section{}
% \label{app:theorem}

% % Note: in this sample, the section number is hard-coded in. Following
% % proper LaTeX conventions, it should properly be coded as a reference:

% %In this appendix we prove the following theorem from
% %Section~\ref{sec:textree-generalization}:

% In this appendix we prove the following theorem from
% Section~6.2:

% \noindent
% {\bf Theorem} {\it Let $u,v,w$ be discrete variables such that $v, w$ do
% not co-occur with $u$ (i.e., $u\neq0\;\Rightarrow \;v=w=0$ in a given
% dataset $\dataset$). Let $N_{v0},N_{w0}$ be the number of data points for
% which $v=0, w=0$ respectively, and let $I_{uv},I_{uw}$ be the
% respective empirical mutual information values based on the sample
% $\dataset$. Then
% \[
% 	N_{v0} \;>\; N_{w0}\;\;\Rightarrow\;\;I_{uv} \;\leq\;I_{uw}
% \]
% with equality only if $u$ is identically 0.} \hfill\BlackBox

% \section{}

% \noindent
% {\bf Proof}. We use the notation:
% \[
% P_v(i) \;=\;\frac{N_v^i}{N},\;\;\;i \neq 0;\;\;\;
% P_{v0}\;\equiv\;P_v(0)\; = \;1 - \sum_{i\neq 0}P_v(i).
% \]
% These values represent the (empirical) probabilities of $v$
% taking value $i\neq 0$ and 0 respectively.  Entropies will be denoted
% by $H$. We aim to show that $\fracpartial{I_{uv}}{P_{v0}} < 0$....\\

\end{document}